\documentclass{article}

\usepackage{amsmath,graphicx,mlspconf}

\usepackage{paralist}
\usepackage{nicefrac}
\usepackage{amsthm}
\usepackage{amsfonts}
\usepackage{amssymb}
\usepackage{xcolor}
\usepackage{hyperref}
\usepackage{siunitx}

\newcommand*{\Scale}[2][4]{\scalebox{#1}{$#2$}}%
\copyrightnotice{979-8-3503-2411-2/23/\$31.00 {\copyright}2023 IEEE}

\toappear{2023 IEEE International Workshop on Machine Learning for Signal Processing, Sept.\ 17--20, 2023, Rome, Italy}

\title{A Flow Artist for High-Dimensional Cellular Data}
\name{\parbox{\linewidth}{\centering%
    Kincaid MacDonald \textsuperscript{1\dag}%
    \qquad Dhananjay Bhaskar \textsuperscript{2,3\dag}%
    \qquad Guy Thampakkul \textsuperscript{4}%
    \qquad Nhi Nguyen \textsuperscript{5} \\%
    \qquad \textit{Joia Zhang} \textsuperscript{6}%
    \qquad \textit{Michael Perlmutter} \textsuperscript{7}%
   \qquad \textit{Ian Adelstein} \textsuperscript{1}%
   \qquad \textit{Smita Krishnaswamy} \textsuperscript{2,3,5,8\ddag}% 
   \thanks{\vspace{-0.25cm} \\ \dag Equal Contribution\\ \ddag Corresponding Author (\texttt{smita.krishnaswamy@yale.edu})}%
   }%
}
\address{%
    \normalsize{\textsuperscript{1}Department of Mathematics, Yale University}\quad%
    \normalsize{\textsuperscript{2}Department of Genetics, Yale School of Medicine}\\%
    \normalsize{\textsuperscript{3}Department of Computer Science, Yale University}\quad%
    \normalsize{\textsuperscript{4}Department of Mathematics, Pomona College}\\%
    \normalsize{\textsuperscript{5}Applied Mathematics Program, Yale University}\quad%
    \normalsize{\textsuperscript{6}Department of Statistics, University of Washington}\\%
    \normalsize{\textsuperscript{7}Department of Mathematics, Boise State University}\quad%
    \normalsize{\textsuperscript{8}Computational Biology and Bioinformatics Program, Yale University}%
}

\begin{document}

%\ninept
\maketitle

\begin{abstract}
We consider the problem of embedding point cloud data sampled from an underlying manifold with an associated flow or velocity. Such data arises in many contexts where static snapshots of dynamic entities are measured, including in high-throughput biology such as single-cell transcriptomics. Existing embedding techniques either do not utilize velocity information or embed the coordinates and velocities independently, i.e., they either impose velocities on top of an existing point embedding or embed points within a prescribed vector field. Here we present \textit{FlowArtist}, a neural network that embeds points while jointly learning a vector field around the points. The combination allows FlowArtist to better separate and visualize velocity-informed structures. Our results, on toy datasets and single-cell RNA velocity data, illustrate the value of utilizing coordinate and velocity information in tandem for embedding and visualizing high-dimensional data.
\end{abstract}
\section{Introduction}
\label{sec:intro}
Many datasets consist of static snapshots of underlying dynamic processes. Examples include single-cell data capturing snapshots of cells in development, molecular measurements of single folds in dynamic folding processes, and financial data measuring static instances of the stock market. These datasets are often embedded using standard dimensionality reduction techniques, which assume that the data satisfies the \textit{Manifold Hypothesis} and aim to find a low-dimensional representation of the resulting manifold of evolving states. But in many cases, additional information about the underlying trajectories of data is available: we know (or can infer) not just the points, but also their velocities. Such data is commonly encountered in biology and biomedical applications, e.g. directed motion and diffusion of water molecules in diffusion-based imaging~\cite{odonnell_introduction_2011}, collective motion of active matter and cells~\cite{gompper_2020_2020}, and transitions between states in gene networks~\cite{bergen_generalizing_2020}. Intuitively, leveraging this flow information should enable a more faithful low-dimensional representation of the manifold — one which captures not only the manifold's geometry but also the underlying dynamics of the manifold's flow.

Embedding data manifolds with flow is a relatively unexplored problem. Most existing solutions fall into two categories: they either first embed the manifold and then draw arrows on top of this embedding, or pre-specify a vector field and embed the manifold within it. For instance, visualizations in the popular single-cell transcriptomics library scVelo perform U-Map or t-SNE to create a point embedding, and overlay arrows at each point such that the arrows point towards the same sets of points as in the ambient space \cite{bergen_generalizing_2020}. Perrault and Joncas perform a similar operation with diffusion maps, performing a projection of the velocities into the diffusion space \cite{perrault-joncas_estimating_2014}. Other techniques, developed for the visualization of directed graphs, use the eigenvectors of the magnetic Laplacian to embed nodes into a prescribed circular flow field \cite{fanuel_magnetic_2018}. 

In this paper, we specifically consider point-velocity pairs sampled from the image of a vector field on a manifold. The vectors may represent velocities of individual points, capturing the evolution of the data through some state space (as with single-cell transcriptomics data), or they may represent a flow field occurring on top of the manifold. Regardless, the additional information provided by the vectors should inform embeddings of the data. If points have similar coordinates but different velocities, they should be embedded at a greater distance than points that share both coordinates and velocities. Moreover, just as more standard manifold learning algorithms embed points into low dimensions, here we desire an equivalent low-dimensional embedding of the velocities.

\section{Background}

\subsection{Manifold learning}
Modern datasets often consist of high dimensional data which lies on a lower dimensional structure such as a linear subspace, a Riemannian manifold, or a collection of such manifolds.  Popular linear dimensionality reduction methods such as Principle Component Analysis seek to find linear relationships in the data and have been widely applied to settings where the data is concentrated near a low-dimensional linear subspace. Analogously, manifold learning algorithms \cite{ coifman_diffusion_2006, moon_visualizing_2019}, which aim to find non-linear structure in the data,  have been applied to datasets concentrated near a low-dimensional Riemannian manifold. Such datasets commonly arise in high-throughput biology such as  single-cell transcriptomics.

However, while dimensionality reduction methods can be used to denoise data, understand the primary modes of variation, and expedite data processing, one often wishes to be able to visualize the data. This is inherently challenging if the intrinsic dimension of the data is greater than two (or possibly three), as it involves sacrificing parts of the data's structure to learn a compressed form suitable for visual comprehension. Methods such as t-SNE \cite{JMLR:v9:vandermaaten08a}, U-Map \cite{mcinnes_umap_2020}, and PHATE \cite{moon_visualizing_2019} have risen to this challenge, aiming to reduce the data to exactly two dimensions while emphasizing the most salient features. These methods differ in which features they consider most salient. For instance, t-SNE emphasizes local neighborhood preservation, producing visualizations with tightly clustered subpopulations, whereas PHATE modifies the diffusion maps algorithm, a popular tool for non-linear dimensionality reduction, to produce a two-dimensional output that preserves smooth trajectories and maintains global geometric features as well as local neighborhoods.

\subsection{Diffusion-based Methods}

Manifold learning algorithms contend with a basic challenge: in the high-dimensional ambient space, the Euclidean distances between the data points are only meaningful locally. To learn the structure of the manifold, many techniques (like Diffusion Maps, and PHATE) apply a kernel to transform the data from point cloud to graph; each data point becomes a node, connected to points that are nearby in Euclidean space. One can use the techniques of geometric learning to “integrate” this local connectivity into global geometric features.

One powerful tool for this analysis is the graph diffusion matrix, $P$, a row-normalized affinity matrix that can be viewed as the transition probabilities of a random walk on the graph. If $A$ is the graph's affinity matrix, and $D$ is the diagonal degree matrix with nonzero entries $D_{ii} = \sum_{j} A_{ij}$ are the row sums of $A$, we define
\begin{equation}
\label{bg:diffusion matrix}
P=D^{-1}A.
\end{equation}
Powering the diffusion matrix $P^t$ yields $t$-step random walk probabilities from each node to every other node.

The diffusion matrix gives rise to the \textit{diffusion map} \cite{coifman_diffusion_2006} (at time 1), which constructs a set of eigen-coordinates from $P$:
\begin{equation}
\label{bg:diffusion map}
x_i \mapsto \Phi(x_i) = [\lambda_0\phi_0(i), \lambda_1\phi_1(i), \dots, \lambda_m\phi_m(i)],
\end{equation}
where $\phi_k$ is the $k^{th}$ right eigenvector of $P$ and $\lambda_k$ is the associated eigenvalue. Coifman and Lafon then prove that Euclidean distances between $\Phi(x_i)$ and $\Phi(x_j)$ approximate the diffusion distances between $x_i$ and $x_j$ on the manifold (i.e., the distance between the probability distribution of a random walker started at $x_i$ and one started at $x_j$)\cite{coifman_diffusion_2006}.

\section{Methods}

\subsection{Problem Setting: A Dual Optimization}\label{sec: problem setting}

We consider the problem of embedding points and velocity from a manifold endowed with a vector field. More specifically, we let $\mathcal{M}$ be a Riemannian manifold embedded in $\mathbb{R}^n$ and let $p_i=(x_i,v_i)\in \mathbb{R}^n\times\mathbb{R}^n$, $1\leq i\leq N$, be a collection of points sampled
from the tangent bundle $T\mathcal{M}$. Furthermore, we assume that there exists a vector field $F:\mathcal{M}\rightarrow T\mathcal{M}$ such that each $(x_i,v_i)$ lies upon the image of $F$. 
 We will think of our data as modeling the trajectory of a particle and, in particular, we will think of $x_i$ and $v_i$ as the position and velocity of the particle at the $i$-th time point.

Our goal is to learn an embedding and a vector field within our embedding space that recreates the dynamics of the vector field $F$. Specifically, given our collection of position-velocity pairs $p_i=(x_i,v_i)$, we wish to learn a point-embedding map $\xi:\mathbb{R}^n \to \mathbb{R}^2$ and a vector field $\psi:\mathbb{R}^2 \to \mathbb{R}^2$ such that

\begin{enumerate}
	\item The embedding approximately preserves the manifold geometry, i.e. the distances of the $x_i$ on the manifold $\mathcal{M}$ are reflected by Euclidean distances in the embedding space.
	\item The embedding preserves the \emph{flow} of the vector field $F$. In other words, when moving along the vector field in the embedding space, we wish to follow a similar trajectory to the one we would moving along the original vector field $F$.
\end{enumerate}

To address these goals, we will construct a network with two sub-networks and two loss functions: (i) a \emph{point embedder} network penalized by a loss dedicated to preserving the manifold geometry, and (ii) a \emph{learnable vector field} with a loss dedicated to preserving the flow geometry. We train these networks together, performing a dual optimization that allows the constraints imposed by the flow geometry to inform the point embedding, and vice versa.
(If one wishes, they may impose additional  soft constraints, such as a preference that the vector field produce trajectories with low curvature, by adding terms to the loss function.)

There’s a rich literature on geometry-preserving point embedding. Many of these methods, such as Diffusion Maps \cite{coifman_diffusion_2006} and PHATE \cite{moon_visualizing_2019} 
embed a point cloud $\{x_i\}_{i=1}^N\subseteq\mathbb{R}^n$ by first using a symmetric kernel $K(\cdot,\cdot)$ to construct an undirected graph whose weighted adjacency matrix satisfies $A_{i,j}=K(x_i,x_j)$. Typically, $K(x_i,x_j)$ is a decreasing function of $\|x_i-x_j\|_2$, and therefore, this graph retains local connectivity information which can be incorporated into a manifold-faithful global representation.
%embed a point cloud by first capturing its structure in a graph of pairwise similarities, which retains local connectivity information which can be integrated into a manifold-faithful global representation. 
To capture the structure of the flows on $\mathcal{M}$, we take a similar tact. We introduce a novel, asymmetric \emph{flashlight kernel} to construct a directed graph from the data points $\{p_i\}_{i=1}^N=\{(x_i,v_i)\}_{i=1}^N$. From this graph, we construct a directed diffusion matrix and an associated directed diffusion map, which we use as the base of our point embedding. This diffusion matrix and diffusion map will form the basis of our penalties for our point embedder $\xi$ and our learnable vector field $\psi$.

\subsection{The Flashlight Kernel and Flow Neighbors}

\begin{figure*}[!h]
\begin{center}
\includegraphics[width=0.85\linewidth]{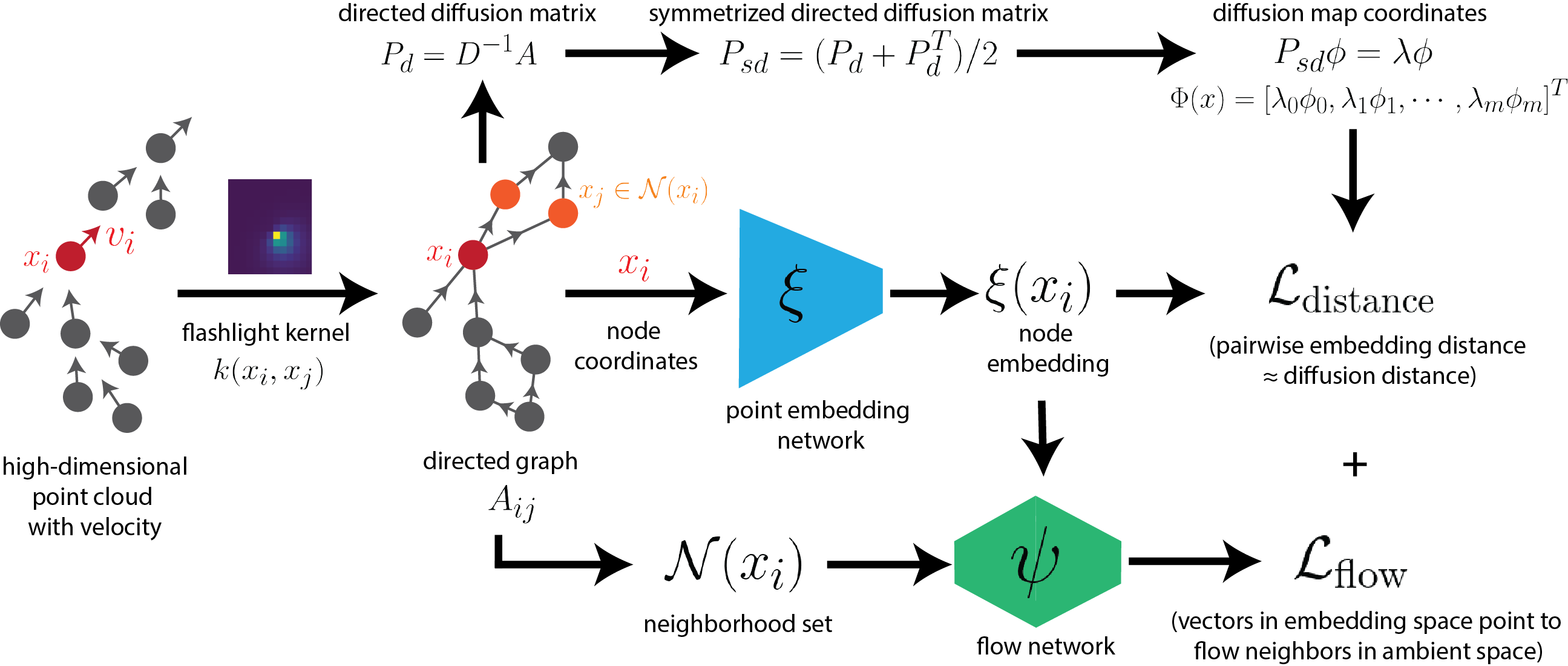}
\end{center}
\vspace{-0.7cm}
\caption{Schematic of loss computation in FlowArtist.}
\label{fig:schematic}
\end{figure*}

In many affinity-based graphs, points $x_i$ are connected based on their proximity in Euclidean space, and the edge weight $A_{i,j}$ is large if $\|x_i-x_j\|_2$ is small. In our setting, we consider position-velocity pairs $p_i=(x_i,v_i)$ and aim to account for both proximity in space and for the flow of the vector field. Therefore, there should be an edge from $x_i$ to $x_j$ only if (i) $x_i$ is close to $x_j$ and (ii) the vector $x_j-x_i$ has a similar direction to $v_i$. This motivates us to introduce a novel kernel $K$ which we name the Flashlight Kernel:
\begin{equation}
\label{eqn:flashlight}
\Scale[0.98]{K(p_i, p_j) = \exp \left(\frac{-\Big[\|x_i-x_j\|_2^2 + \beta \left( \|v_i\|_2 - \left\langle v_i, \frac{x_j - x_i}{\|x_j - x_i\|_2}\right\rangle \right) \Big]}{\sigma}\right).}
\end{equation}
We then define a directed adjacency matrix $A$ by $A_{i,j}=K(p_i,p_j)$. 
%
%\begin{figure}
%  \begin{center}
%    \includegraphics[width=0.5\linewidth]%{Figures/flashlight-kernel.png}
%  \end{center}
%  \caption{The Flashlight Kernel on a plane: affinities from the origin, with a velocity towards the bottom right.}
%  \label{fig:flashlight-kernel}
%\end{figure}
%
This combines the standard Gaussian kernel with a flow-affinity term that measures the correspondence between the velocity at $x_{i}$ and the direction towards its neighbor $x_{j}$. Notably, in the case where the unit vector $\frac{v_i}{\|v_i\|_2}$ is equal to $\frac{x_i-x_j}{\|x_i-x_j\|_2}$ then the second term is zero and $K(\cdot,\cdot)$ reduces to a standard Gaussian kernel. (See Figure \ref{fig:schematic} for a visual depiction of the flashlight kernel.)

We note that \cite{coifman_diffusion_2006} also considered a kernel for points in the tangent bundle. Our flashlight kernel differs from the one introduced in \cite{coifman_diffusion_2006} in the treatment of points that go backwards to the velocity, i.e., when $\frac{v_i}{\|v_i\|_2}\approx - \frac{x_i-x_j}{\|x_i-x_j\|_2}$. The flashlight kernel evaluates to zero in this case (i.e., backwards motion is impossible), while the kernel used in \cite{coifman_diffusion_2006} treats this setting as if $\frac{v_i}{\|v_i\|_2}\approx \frac{x_i-x_j}{\|x_i-x_j\|_2}$ (i.e. backwards motion is treated the same as forward motion, but motion \textit{perpendicular} to the flow is impossible).
Using the flashlight kernel, one can define the \emph{flow neighborhood}, $\mathcal{N}(x_i)$, of a point $p_{i}=(x_i,v_i)$ as the set of $k$ points $p_j$ such that $K(p_i,p_j)$ is as large as possible. Our flow-geometry loss will aim to preserve these neighborhoods.

\subsection{The Directed Diffusion Matrix and Directed Diffusion Map}

\begin{figure}
  \begin{center}
    \includegraphics[width=0.9\linewidth]{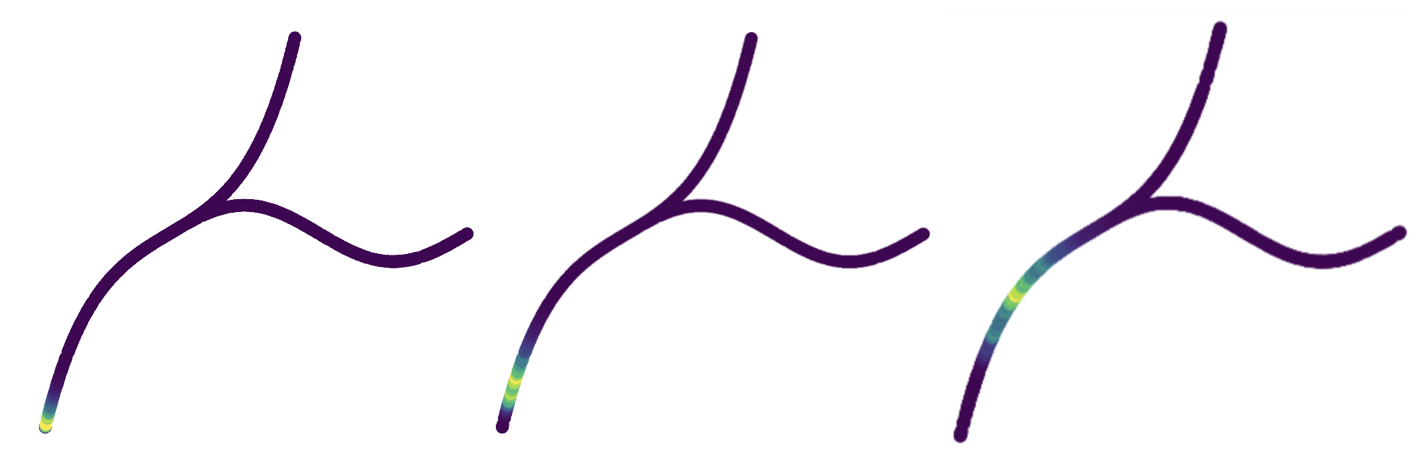}
  \end{center}
  \vspace{-0.7cm}
  \caption{The diffusion probabilities from the lowest point on the branch, at different scales of diffusion.}
  \label{fig:directed-diffusion}
\end{figure}

\begin{figure}[h]
  \begin{center}
    \includegraphics[width=0.75\linewidth]{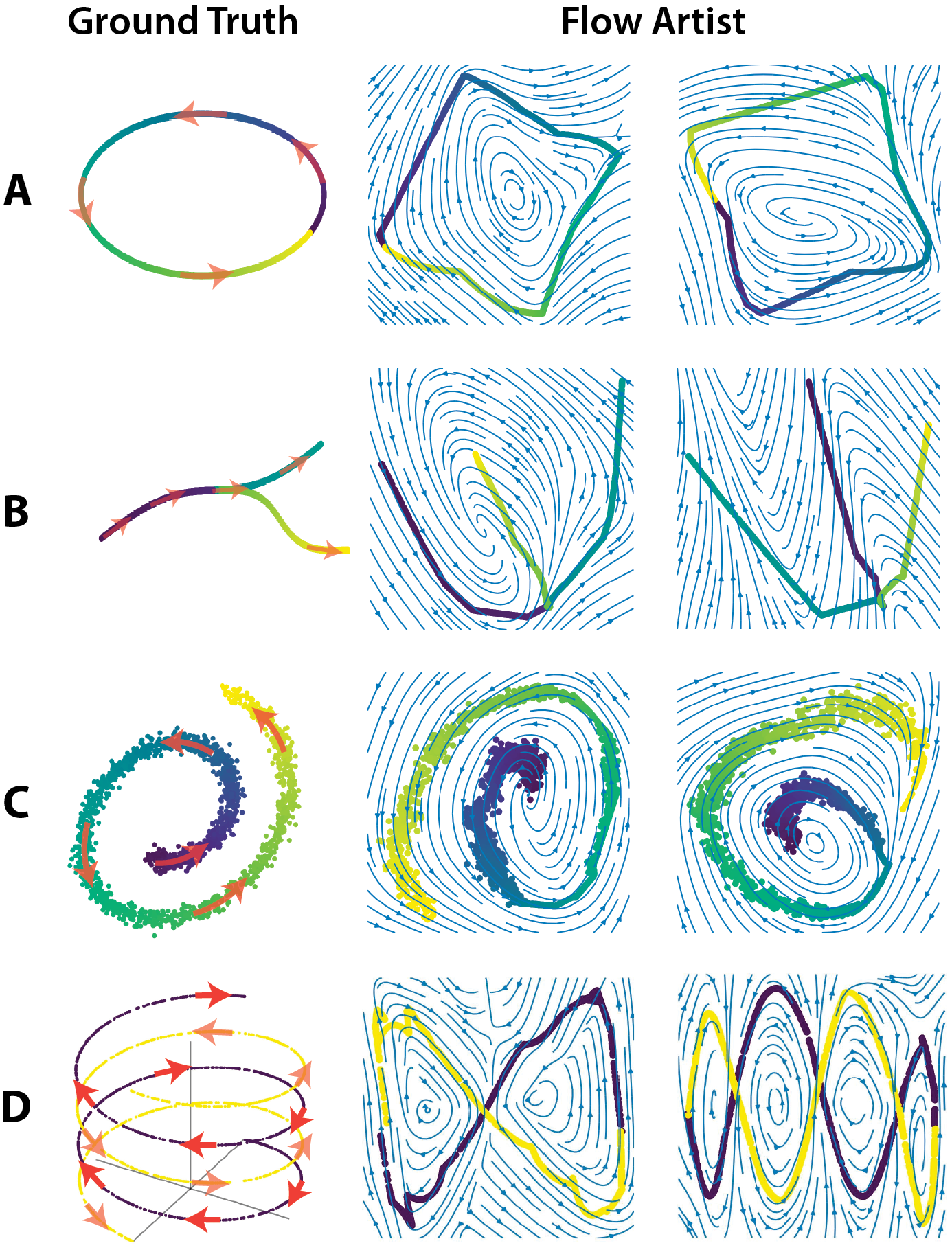}
  \end{center}
  \vspace{-0.7cm}
  \caption{Embedding of synthetic datasets using FlowArtist.}
  \label{fig:toy-dataset}
\end{figure}

Using the flashlight kernel, we can construct a directed graph from our data points $\{p_i\}_{i=1}^N$. We can now apply some of the traditional methods of graph analysis, beginning with the diffusion matrix. Let $A$ be the flow-affinity matrix constructed with the flashlight kernel, and let $D$ be the diagonal matrix whose nonzero entries $D_{ii} = \sum_{j}A_{ij}$ are the row sums of $A$.\footnote{Note that since $A$ is asymmetric, $D$ is the out-degree matrix of the graph corresponding to $A$.} We then define the directed diffusion matrix as the row-normalized affinities $P_d = D^{-1}A$.

The directed diffusion matrix has several interesting properties. Similar to a diffusion matrix resulting from a symmetric kernel, it can be viewed as the transition probabilities of a random walk on the graph and by powering the matrix $P^t$, we obtain the transition probabilities of a $t$-step random walk. But while the random walk on an undirected graph remains based around the center, growing increasingly diffuse with time, the random walk modeled by directed diffusion retains its concentration while traveling along the manifold in the direction of flow as illustrated in Figure \ref{fig:directed-diffusion}. One can show that the degree to which the concentration of the diffusion is retained depends on the determinant of $P$; a high determinant means that the $t$-step random walk will be less diffuse, and hence that the underlying vector field is more homogenous.

In order to be able to compute eigenvectors and apply traditional diffusion maps to our data, we also consider the symmetrized  diffusion matrix $P_{sd} = \frac{1}{2}(P_{d} + P_{d}^T)$. In $P_{sd}$, two points are connected only if one can flow to the other with a single step of diffusion. %Points which are mutually unreachable by flow are disconnected (even if connected in a distance-based graph).
Notably, $P_{sd}$ is not the same as the matrix obtained by first symmetrizing $A$ and then normalizing by the resulting degree matrix. 
This enables the directed diffusion map to create a flow-informed manifold embedding, in which points inaccessible by flow are placed further apart than in a traditional diffusion map. 

\subsection{Preserving Manifold Geometry with the Point Embedder}

We penalize our network by two loss functions corresponding to the goals outlined in Section \ref{sec: problem setting}. Our first loss function aims to ensure that we learn an embedding $\xi:\mathbb{R}^n \to \mathbb{R}^2$  such that the Euclidean distances between the embedded points $\xi(x_{i})$ approximate the distances between the $x_i$ on the underlying manifold $\mathcal{M}$. Specifically, we define
\begin{equation}
    \label{eqn:Manifold Distance Loss}
       \mathcal{L}_{\text{distance}} = \| D_{\text{embed}} - D_{\text{manifold}}\|^2_{F},
\end{equation}
where $D_{\text{embed}}$ is the matrix of pairwise Euclidean distances between the $\xi(x_{i})$ (in $\mathbb{R}^2)$ and $D_{\text{manifold}}$ is a matrix of approximate distances between the $x_{i}$ on the manifold $\mathcal{M}$. In our experiments, we use the diffusion distance on the manifold approximated as in \eqref{bg:diffusion map} (where here the $\phi_k$ and $\lambda_k$ are the eigenvectors and eigenvalues of $P_{sd}$).  Using the directed diffusion map adds additional information from the manifold’s flow, enabling easier satisfaction of our second loss.

\subsection{Preserving Flow Geometry}
Our second loss function aims to ensure  that our learnable vector field $\psi:\mathbb{R}^2 \to \mathbb{R}^2$ properly recreates the flow of the true vector field on $\mathcal{M}$ in the sense that if $K(p_i,p_j)$ is large then $\psi(x_i)\approx \xi(x_j)-\xi(x_i)$.
%
%To motivate our learnable vector field to recreate the flow geometry, we penalize the velocity at each embedded point $\xi(x_{i})$ to point towards the same points as in ambient space. 
This motivates us to define the  \emph{Flow Neighbor Loss} by
\begin{equation}
  \label{eqn:Flow Loss}
    \mathcal{L}_{\text{flow}} = \sum\limits_{x_j \in \mathcal{N}(x_{i})} \| (\xi(x_j) - \xi(x_i)) - \psi(x_i)\|_2^2
\end{equation}
where $\mathcal{N}(x_{i})$ is the set of $k$ points with highest flow-affinity to $x_{i}$, $\xi(x_i)$ is the embedding of $x_{i}$, and $\psi(x_i)$ is the learned vector field at $x_i$.

We note that there are many other possible choices for our second loss function. For example, one could apply the flashlight kernel to the embedded points and velocities, and treat the resulting probabilities as predictions of the flow neighbors, penalized by a contrastive loss. In practice, we find the simpler formulation of the Flow Neighbor Loss achieves equivalent results and is computationally much faster. 
Additionally, one can regularize the embedding with extra terms that act directly on the learned vector field $\psi$. One such penalty is a Laplacian smoothness regularization, $sum_i\frac{v_i^tLv_i}{v_i^tv_i}$, where $v_i$ the column vector containing the $i^{th}$ output coordinates of $\phi$ applied to each $x \in X$ and $L$ is the graph Laplacian derived from $X$.\footnote{The effect of enforcing smoothness depends on the smoothness of the underlying data. For example, when embedding a Swiss Roll with FlowArtist, we can force the embedding to "unroll" with this smoothness regularization. But in more complex datasets, like our double helix, smoothness impedes the embedding; abrupt changes in direction are needed for the low-dimensional vector space to recreate the higher dimensional phenomena.} 

Given $\mathcal{L}_{\text{flow}}$ and $\mathcal{L}_{\text{distance}}$,
we then train FlowArtist to minimize the total loss, $\mathcal{L} := \mathcal{L}_{\text{flow}} + \mathcal{L}_{\text{distance}}$. The result is an embedding which, though partially informed by estimates of $D_{\text{manifold}}$, also aims to preserve the manifold’s flow. In our experiments,
we parameterize $\xi$ and $\psi$ as multi-layer perceptrons (each under 10 layers), with Leaky ReLU activations, and optimize them against our loss with stochastic gradient descent implemented by the Adam optimizer, with a learning rate of $10^{-3}$. The compact size of our model allows it to train quickly on CPU, where it can embed each of our toy datasets in under five minutes. Because our losses rely on having points from within the same neighborhood, we perform batching pointwise: each batch contains a central point, a smattering of its neighbors, and a random selection of non-neighbors. This allows the model to attend to both local and global geometry with each optimization step.
% TODO: Citation for Adam optimizer

Although we implemented the point-embedder and learnable vector-field as simple feed-forward neural networks, each could in principle be replaced by a more specialized module, optimized by our loss functions, that could extend and improve the capabilities of FlowArtist. For instance, to create FlowArtist embeddings of graphs without associated positional data, one could use a graph autoencoder that learns positional node embddings, such as the one described in Satorras's \textit{E(n) GNN}, combined with a learnable vector field and our Flow loss. Likewise, our learnable vector field could be made more expressive by use of a dedicated network, such as VectorNet.
% TODO: find citation for neural vector fields nnets
(although in our experiments, we found this particular network more temperamental to train than the simpler feed-forward network.)

\section{Results}

\begin{figure}[h]
  \begin{center}
    \includegraphics[width=0.8\linewidth]{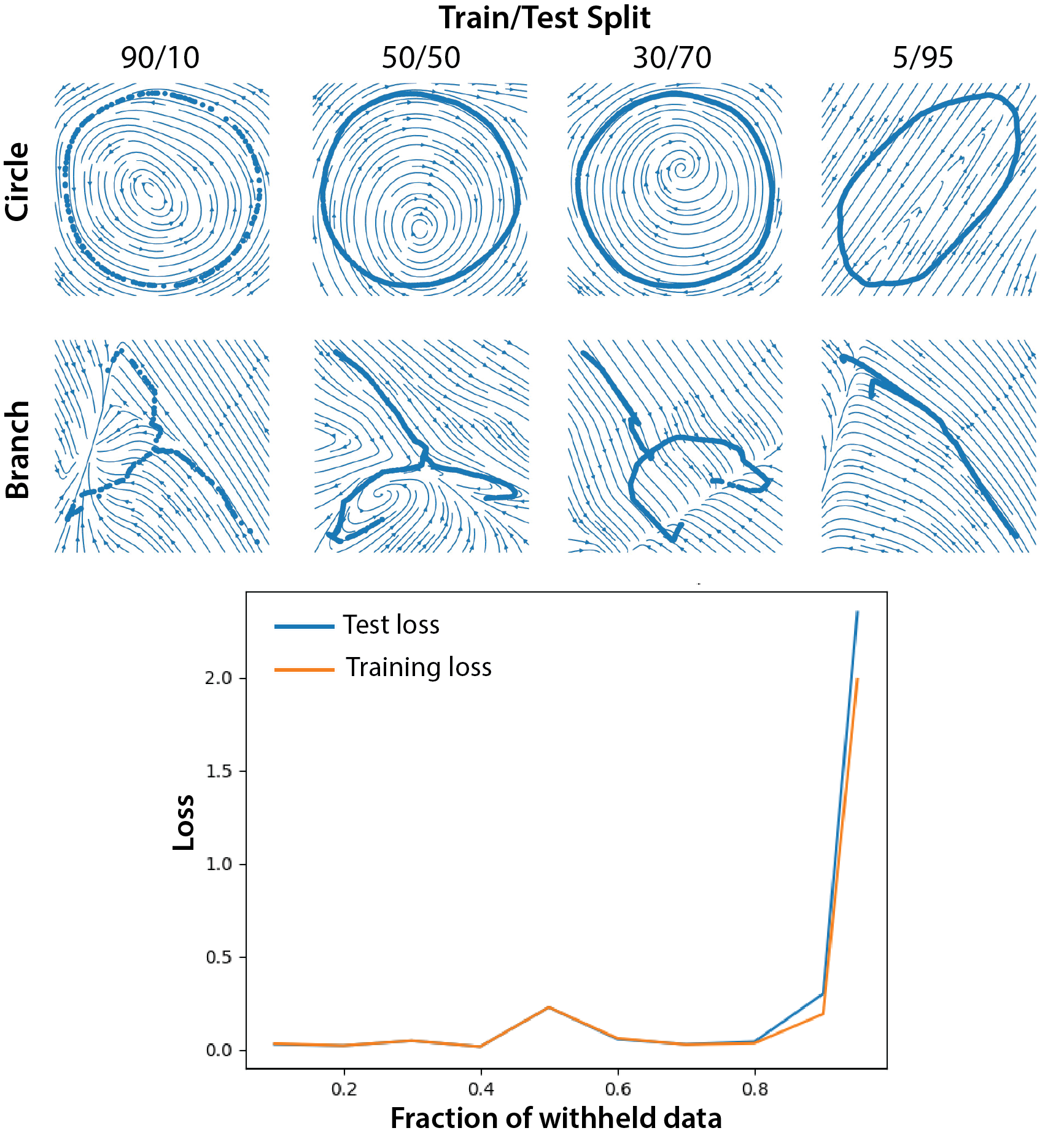}
  \end{center}
  \vspace{-0.7cm}
  \caption{Vector field visualization and loss values (branching data) at different train/test splits.}
  \label{fig:traintest}
\end{figure}

We demonstrate the utility of FlowArtist on both toy datasets and simulated single-cell datasets.\footnote{Code available at: \href{https://github.com/KrishnaswamyLab/FlowArtist}{https://anonymous.4open.science/r/FlowArtist/}} First, we test FlowArtist on toy datasets consisting of points arranged in the shape of a circle, tree branch, spiral, and double helix with constant magnitude velocity vectors, indicated by red arrows, as shown in Fig.~\ref{fig:toy-dataset}. The double helix (Fig.~\ref{fig:toy-dataset}D) consists of two disjointed helices (colored in indigo and yellow) with velocities in opposite directions. Point embeddings and velocity vector fields generated using two random initializations of the FlowArtist network are shown in Fig.~\ref{fig:toy-dataset}. In all cases, the embedding of points preserves the geometry of the input data and the predicted velocity vectors (visualized as streamlines) agree with the ground truth. The quality of the point embedding and predicted vector field demonstrated robustness across various train/test splits (Fig.~\ref{fig:traintest}). Our experiments with synthetic data confirm the stability of the learning algorithm's performance, irrespective of the data partitioning strategy.

\begin{figure}
  \begin{center}
    \includegraphics[width=0.95\linewidth]{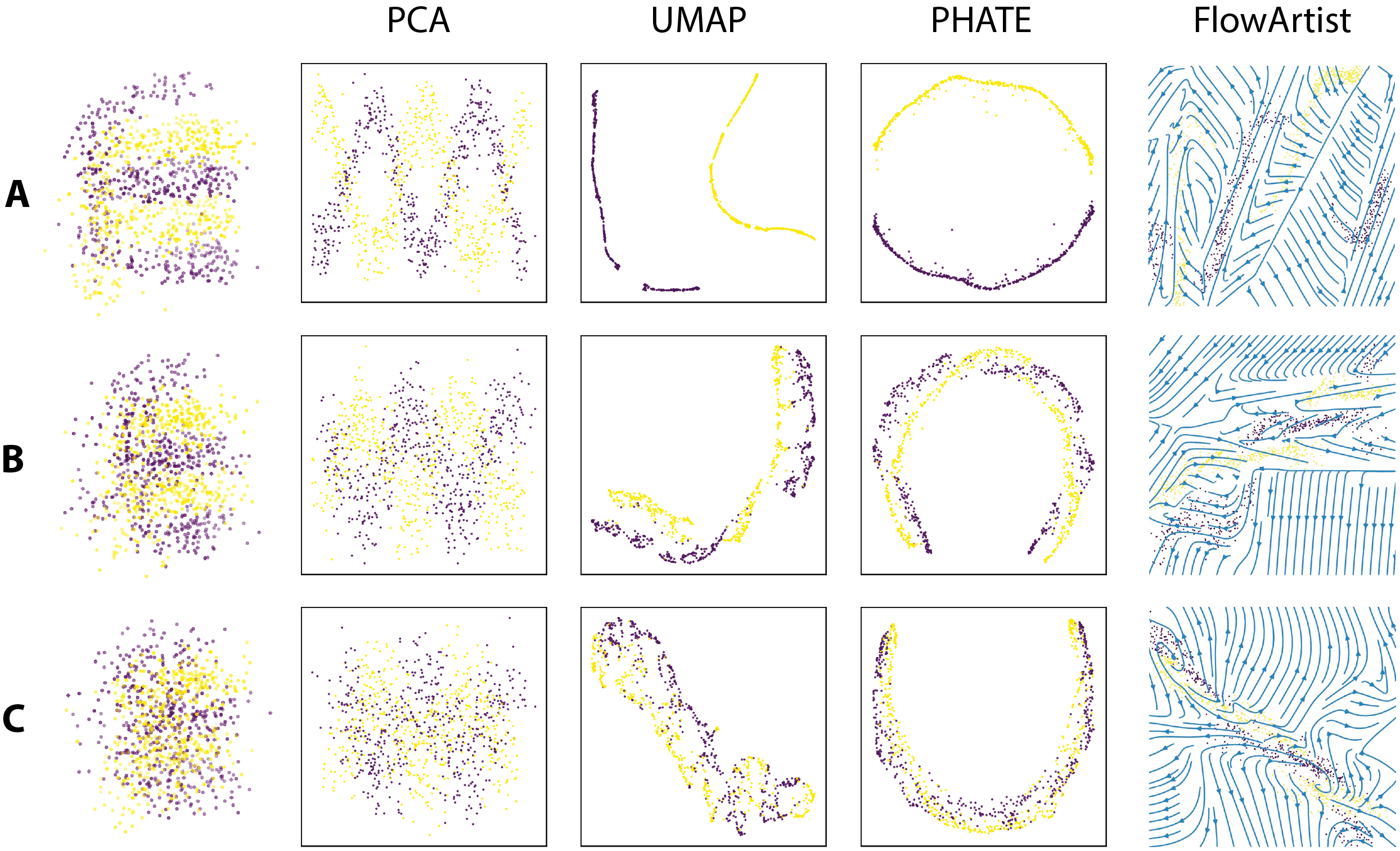}
  \end{center}
  \vspace{-0.7cm}
  \caption{Embeddings of the double helix dataset with varying levels of additive Gaussian noise. (A) $\mathcal{N}(0,0.25)$, (B) $\mathcal{N}(0,0.5)$, (C) $\mathcal{N}(0,0.75)$. }
  \label{fig:noisy-double-helix}
\end{figure}

% flow critical datasets
With the addition of noise, the double helix becomes a particularly salient example of FlowArtist's advantages against more traditional manifold-learning algorithms. If the level of noise is sufficiently high, the two strands of the double helix merge into a homogeneous cloud, rendering the dataset indecipherable (Fig. \ref{fig:noisy-double-helix}). Under PCA, the yellow and the purple strand overlap for all three noise levels, $.25, .5$, and $.75$ and the strands become increasingly intertwined as the noise level increases. At noise level $.25,$ UMAP and PHATE produce disjoint embeddings of the two strands. However, the strands overlap at noise levels $.5$ and $.75$. The point embedder of FlowArtist faces a similar issue, particularly at noise level $.75$. However, the strands can be distinguished by their velocities because the yellow points which are near purple points have arrows pointing in opposite directions. Thus the FlowArtist produces a representation which allows one to distinguish the individual helices.

\begin{figure}[h]
  \begin{center}
    \includegraphics[width=0.95\linewidth]{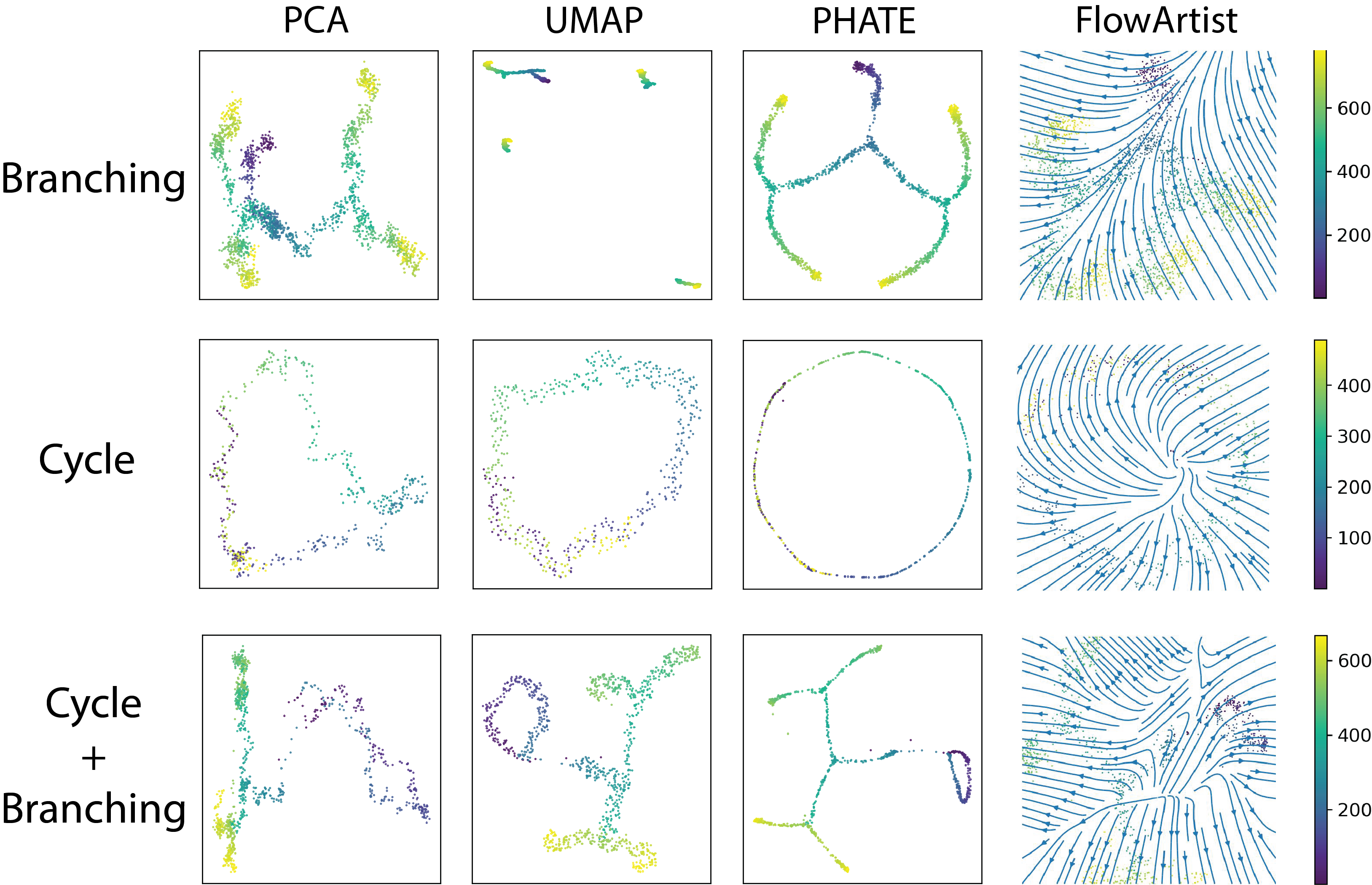}
  \end{center}
  \vspace{-0.7cm}
  \caption{Embeddings of scRNA-seq datasets generated via VeloSim \cite{zhang2021velosim} to simulate cell cycle and differentiation.}
  \label{fig:VeloSim}
\end{figure}

Next, we consider simulated single cell gene expression datasets generated by VeloSim \cite{zhang2021velosim}, a recently introduced software package that models dynamics of a gene regulatory network specified by the user. VeloSim generates unspliced and spliced RNA counts for each gene which are used estimate the RNA velocity. We simulated cell differentiation and cycling processes in $2000$ cells and $500$ genes using the simulator. A more complex process where cells cycle before branching was also simulated. We visualized the data in 2D colored by pseudotime using various dimension reduction techniques (Fig. \ref{fig:VeloSim}). We trained FlowArtist using the simulation dataset to obtain a low-dimensional embedding of the cells and the RNA velocity. FlowArtist preserved the geometry of the branching dataset and the generated velocity vectors that agreed with pseudotime. Although, FlowArtist did not fully capture the periodicity in the cell cycle dataset, it generated an outward spiral with the arms of the spiral correctly pointing in the direction of increasing pseudotime. Finally, in the dataset consisting of a cycle followed by a branching trajectory, FlowArtist correctly predicted the branching process but did not capture initial the cell cycle. This can be addressed by training FlowArtist to predict velocities at a finer resolution, albeit with more computational expense. Future work will aim to address this limitation by adapting the resolution of the predicted vector field to match the density of the data.

\section{Conclusion}

We have introduced FlowArtist, a novel method for embedding and visualizing data sampled from a vector field defined on an unknown manifold $\mathcal{M}$. We have validated our method on both toy datasets such as a tree branch and a double helix and on synthetic RNAseq datasets generated by VeloSim. These results show that FlowArtist is effectively able to capture both the underlying geometry of the manifold and the flow of the vector field.

The key aspects of FlowArtist are a novel flashlight kernel which allows us to represent the data by a directed graph and a two-part loss function which makes sure that the embedding preserves both the geometry of the manifold and the flow of the vector field. Unlike most existing techniques which first embed the points and then define a vector field or which first define a vector field and then embed the points, FlowArtist learns the point embedding and the vector field jointly. Lastly, we note that while in this paper we have focused on data visualization, our methods could be readily adapted to embeddings into higher dimensions and our flashlight kernel may be used for other tasks in which one wants to represent vector fields by directed graphs. We leave further  exploration of these ideas to future work.

\bibliographystyle{IEEEbib}
\bibliography{refs}

\end{document}